\documentclass[runningheads]{llncs}
\usepackage{graphicx}
\usepackage{amsmath,amssymb} % define this before the line numbering.
\usepackage{color}
\usepackage[width=122mm,left=12mm,paperwidth=146mm,height=193mm,top=12mm,paperheight=217mm]{geometry}

\begin{document}
% \linenumbers
\pagestyle{headings}
\mainmatter
\def\ECCV18SubNumber{6}  % Insert your submission number here

\title{3D Human Body Reconstruction from a Single
  Image via Volumetric Regression}

\titlerunning{ECCV-18 submission ID \ECCV18SubNumber}

\authorrunning{ECCV-18 submission ID \ECCV18SubNumber}

% \author{Anonymous ECCV Submission}
\author{Aaron S. Jackson \textsuperscript{1},
  Chris Manafas \textsuperscript{2},
  Georgios Tzimiropoulos \textsuperscript{1}
}
% \institute{Paper ID \ECCV18SubNumber}
\institute{\textsuperscript{1} School of Computer Science, The University of Nottingham,
  Nottingham, UK \\
  \texttt{\{aaron.jackson,yorgos.tzimiropoulos\}@nottingham.ac.uk} \\ [1em]
  \textsuperscript{2} 2B3D, Athens, Greece \\
  \texttt{chris.manafas@2b3dglobal.com}
}

\maketitle

\begin{abstract}
  This paper proposes the use of an end-to-end Convolutional Neural
  Network for direct reconstruction of the 3D geometry of humans via
  volumetric regression. The proposed method does not require the
  fitting of a shape model and can be trained to work from a variety
  of input types, whether it be landmarks, images or segmentation
  masks. Additionally, non-visible parts, either self-occluded or
  otherwise, are still reconstructed, which is not the case with depth
  map regression. We present results that show that our method can
  handle both pose variation and detailed reconstruction given
  appropriate datasets for training.

\end{abstract}

\begin{figure}
  \centering
  \includegraphics[width=0.8\linewidth]{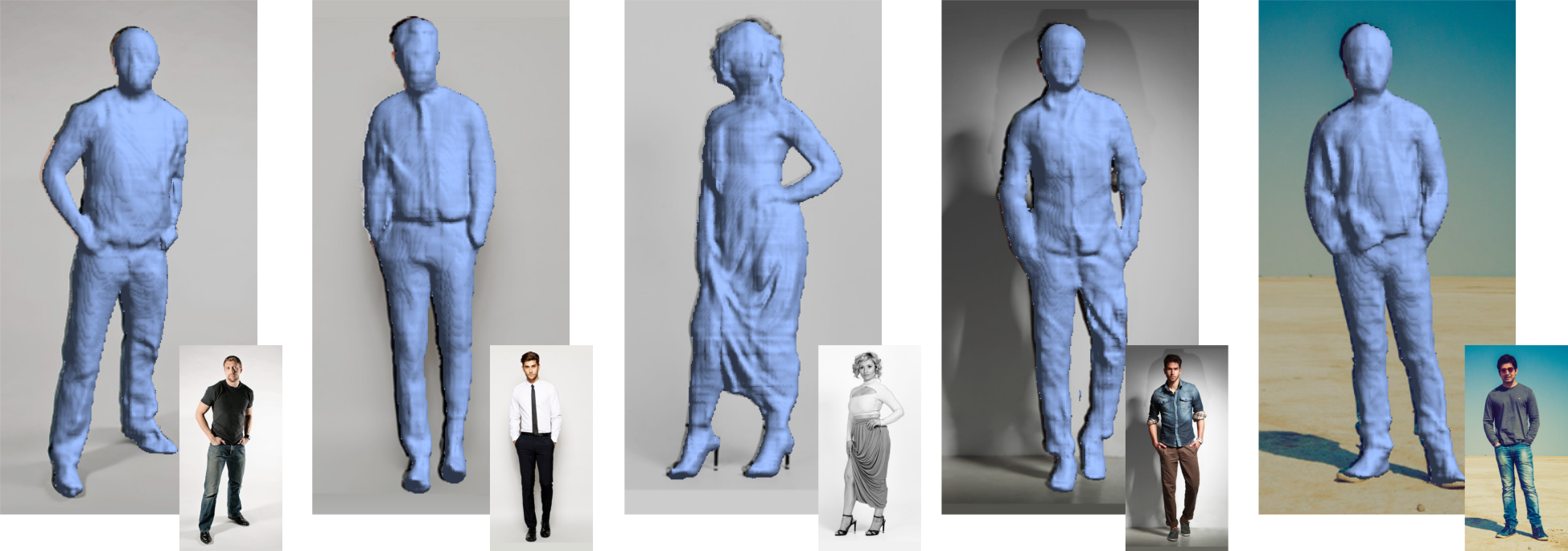}
  \caption{Some example results using our method when trained with a
    high quality detailed training set.}
  \label{fig:topdemo}
\end{figure}

\section{Introduction}

3D reconstruction is the process of estimating the 3D geometry from
one or more 2D images. In this work, we focus on reconstruction of the
human body from a single image, including the non-visible parts which
have been self-occluded. Our method builds upon that
of~\cite{jackson2017vrn} where a 3D face is directly regressed from a
single image, using what they refer to as a ``Volumetric Regression
Network'' (VRN). In this paper, we show that the same idea can be applied to
other deformable objects, in particular, the human body. This poses an
array of challenges which are not present when reconstructing the
face. While we are still only reconstructing an object of a single
class, the body has many more axes of rotation compared to a face. As
such, human body reconstruction is often considered to be a very
difficult problem.

\paragraph{Motivation.} The pipelines required for 3D human
reconstruction (and 3D reconstruction in general) are typically based
on solving difficult non-convex optimisation problems. Perhaps the
most common approach to 3D human body reconstruction is to fit a shape
model. For example, the recent method of \cite{bogo2016smplify}, uses
optimisation to fit a 3D shape model to 2D body joints. However,
optimisation methods are sensitive to initialization and are easily
trapped to local minima, both of which are exacerbated by occlusions
and potential scale changes.

In this work, we aim to significantly reduce the complexity of
standard 3D human reconstruction techniques - to the point where it
could just as easily be treated as a segmentation task. We do this by
directly regressing a volumetric representation of the 3D geometry
using a standard, spatial, CNN architecture, where the regressed
volumetric structure is spatially aligned with the input. Notably, we
do \textbf{not} regress a depth map; the 3D structure is regressed as
slices and recovered from its volumetric representation using a
standard surface extraction algorithm, such as Marching
Cubes~\cite{lorensen1987marching}. In summary, our main contributions
in this work are as follows:

\begin{enumerate}

\item We are the first to apply volumetric regression
  networks~\cite{jackson2017vrn} to the problem to human body
  reconstruction, not just human faces.

\item We propose several improvements to the network architecture
  described in~\cite{jackson2017vrn}, which show significant
  performance improvements. These include introducing intermediate
  supervision, using more advanced residual modules and altering the network
  structure by increasing the number of hourglass modules by reducing
  the number of residual modules.

\item We show that VRN is capable of reconstructing complex poses when
  trained on a suitable dataset.

\item Finally, we show that given high quality training data, our
  method can learn to produce previously unseen, highly detailed, full
  3D reconstructions from only a single image. To the best of our
  knowledge, there is no other method capable of obtaining results
  with such high fidelity and reliability as ours.
\end{enumerate}

The remainder of the paper is structured as follows: First, a review
of closely related work on 3D human body reconstruction and human pose
estimation is given. We then describe our method, including the
volumetric representation we have already mentioned briefly, followed
by the datasets and the training procedure. Next we will discuss
several architectural variants of VRN, followed by results from a
network trained with pose-variant data, but little detail. Finally, we
will show results which have been generated by training a model with
highly detailed data.

% 3DMM fitting
% Depth estimation

% In~\cite{5597194} no fitting is performed. Instead, the closest
% pose from a dataset of many pre-existing reconstructions

\section{Closely Related Work}

In this section we will give an overview of recent and popular
approaches to human pose estimation (often a prerequisite to human
reconstruction) and 3D reconstruction methods, both working from
images and from landmarks.

\paragraph{Human Pose Estimation.} All modern approaches to estimating
the human pose are based on methods employing CNNs. These methods
generally fall into one of two categories. The first is to directly
regress the coordinates of the joints using an L2 (or similar)
loss~\cite{li20143d,park20163d,tekin2016structured,tekin2016direct,zhou2016deep,chen2016synthesizing,ghezelghieh2016learning}.
In particular,~\cite{park20163d} estimate the 3D pose by combining the
2D predictions with image features. An autoencoder is employed
in~\cite{tekin2016structured} to constrain the pose to something
plausible. Similarly,~\cite{zhou2016deep} have the same goal but
achieve this by using a kinematic model. Synthetic data is used for
the full training procedure in~\cite{chen2016synthesizing}, to ensure
that the network is trained with accurate data. However,
in~\cite{ghezelghieh2016learning}, they only augment their existing
training set with synthetic data. The second approach to CNN based
human pose estimation is to regress a
heatmap~\cite{zhou2016sparseness,pavlakos2017coarse,mehta2017vnect,bulat2016human}. In~\cite{zhou2016sparseness}
they do this from video. In~\cite{pavlakos2017coarse} they regress a
3D heatmap, which is a similar idea to our own work. Another temporal
based approach is described in~\cite{mehta2017vnect}, where the 2D
landmarks are first refined also as a heatmap. A part based heatmap
regression approach is shown in~\cite{bulat2016human}.

In this work, we do not aim to estimate the human pose as a set of
coordinates. Instead, we aim to reconstruct the full 3D geometry of
the human, from just a single image. This includes any parts of the
body which are self occluded. However, in doing so, we optionally make
use of information from a human pose estimation step, which is
provided to the network as 16 channels, each with a Guassian centred
above the respective landmark.

\paragraph{Reconstruction from Image.} Many human reconstruction
methods estimate the geometry from one or more images. For
example,~\cite{balan2007detailed,grest2005human,guan2009estimating}
fit a model based on a single RGB or grey scale images. In particular,
\cite{grest2005human} fit a skeleton model to the image by estimating
the scale and pose of each body part
separately. In~\cite{guan2009estimating}, they fit a shape model
initialised by a user clicking on separate body parts, assisted by a
segmentation mask. Another shape model based approach is proposed
in~\cite{balan2007detailed}, using the SCRAPE
model~\cite{anguelov2005scape}, which is fitted with a stochastic
optimisation step. A general shape fitting method for reconstruction
is proposed in~\cite{chen2010inferring}, where two Gaussian models are
used - one for shape and one for pose, by solving non-linear
optimisation problems. The authors demonstrate this method on human
bodies and sharks. In~\cite{jiang20103d}, a single image and
corresponding landmarks are used to lookup a similar human pose using
a kd-tree, containing about 4 million examples. A method intended for
multi-instance model fitting from a single image is described
in~\cite{Zanfir_2018_CVPR}.

Several methods aim to estimate the 3D geometry using only the
landmarks extracted via human pose
estimation~\cite{bogo2016smplify,ramakrishna2012reconstructing}. Particularly,
SMPLify~\cite{bogo2016smplify} (which uses the SMPL
model~\cite{loper2015smpl}), was extended to also include further
guidance from an segmentation mask
in~\cite{varol2017learning}. However, such an approach will never be
able to capable of regressing finer details, unless information from
the image is also captured.

Aside from SCRAPE~\cite{anguelov2005scape} and
SMPL~\cite{loper2015smpl}, mentioned earlier, Dyna, the shape model
capable of capturing large variations in body shape is presented
in~\cite{Dyna:SIGGRAPH:2015}, but without an accompanying fitting
method from a single image. A very recent shape model called Total
Capture~\cite{Joo_2018_CVPR} captures many aspects of the body which
are typically ignored by other shape models, including the face and
hands.

Our work is different from all of the aforementioned methods in that
we do not regress parameters for a shape model, nor do we regress the
vertices directly. Further more, our method skips the model generation
step entirely, which avoids the need to find dense correspondence
between all training examples. Instead, we constrain the problem to
the spatial domain, and directly regress the 3D structure using
spatial convolutions in a CNN, via a volumetric representation from
which the full 3D geometry can be recovered.

\section{Method}

This section describes our proposed method, including the voxelisation
and alignment procedures.

\subsection{Volumetric Regression}

In this work, our goal is to reconstruct the full geometry of a human
body from just a single image.  There are several ways of estimating
the geometry using deep learning. The first is to directly regress the
vertices using a top-down network such as VGG~\cite{simonyan2014very}
or ResNet~\cite{he2016deep} trained with an L2 loss. This has at least
two disadvantages: firstly it requires the training data to be
resampled to have a fixed number of vertices, which implies finding
correspondence between all vertices of all meshes. Secondly, and more
importantly, training a network to directly regress a very large
number of vertices is hard. A common, and more efficient alternative
is to regress the parameters of a 3D shape model. However, as these
parameters are not scaled equally, it is necessary to employ
normalisation methods, such as weighting the outputs using the
Mahalanobis distance which has been also proven challenging to get it
working well~\cite{jackson2017vrn}. Additionally, 3D shape model based
approach are known to be good at capturing the coarse shape but less
able at capturing fine details (in the case of detailed 3D
reconstruction).

To eliminate the aforementioned learning challenges, we reformulate
the problem of 3D reconstruction by constraining it to the spatial
domain, using a standard convolutional neural network. Our approach
can be thought of as a type of image segmentation where the output is
a set of slices capturing the 3D geometry. Hence architecturally one
can use standard architectures for (say, semantic)
segmentation. Following the work of~\cite{jackson2017vrn} on human
faces, we do this by encoding the geometry of the body in a volumetric
representation. In this representation, the 3D space has been
discretised with a fixed dimensionality. Space which is
\textit{inside} the object is encoded as a voxel with value equal to
one. All other space (i.e. background or unknown object classes) are
encoded with a voxel with a value equal to zero. For this particular
application, the dimensionality of our volumes are
$128\times 128\times 128$, which given the level of detail in our
training set, is more than adequate (although we show in
Section~\ref{sec:detailed} results with much greater detail, and only
a slightly larger volume). One of the main advantages of this
representation is that it allows the non-visible (self-occluded or
otherwise) parts of the geometry to also be reconstructed. This is not
the case in methods attempting to reconstruct the body using depth map
regression.

One of the most important aspects to note in the case of training a
volumetric regression network is that the input and output must be
spatially aligned. Put simply, the 2D projection of the target object
should do a reasonable, if not very good, job at segmenting the
input. Through experimentation, we have found that it is possible to
ignore spatial alignment, as long as the pose is fixed (i.e. always
frontal). However, ignoring spatial alignment will severely impact the
performance of the method.

When trained to receive guidance from human pose estimation, landmarks
are passed to the network as separate channels, each containing a
Gaussian centred over the landmark's location. The Guassians have a
diameter of approximately 6 pixels.

\subsection{Dataset and Voxelisation}

While Human3.6M~\cite{IonescuSminchisescu11,h36m_pami} does include
its own 3D scans, they are not in correspondence with the video
frames. As such, we produced our training data by running
SMPLify~\cite{bogo2016smplify} on the Human3.6M dataset. The landmarks
required by SMPLify were generated using the code made available
with~\cite{bulat2016human}. The fitted meshes were voxelised at a
resolution of $128\times 128 \times 128$. In terms of depth, the
meshes are first aligned by the mean of the Z component. However,
through experimentation, we found that as long as the Z alignment is
performed in a seemingly sensible way, and remains stable across all
images, the network will learn to regress the 3D structure without
issue. Random scale augmentation was performed in advance of the
training procedure, as doing this on-the-fly (for 3D volumes) can be
quite demanding in terms of CPU usage.

An unfortunate side effect of using SMPLify to generate our training
data is that it is not possible to regress features such as fingers or
facial expressions. SMPLify does not model these, and as such, their
pose remains fixed across all images. It also becomes a bottleneck in
terms of performance. We show in Section~\ref{sec:detailed}, using a
different dataset, that very high quality reconstruction is also
possible with our proposed method.

\subsection{Training}

Our end-to-end network was trained using
RMSProp~\cite{hinton2012neural} optimisation with a learning rate of
$10^{-4}$, which was reduced after approximately 20 epochs to
$10^{-5}$ for 40 epochs. We did not observe any performance
improvement by reducing this learning rate further. A batch size of 6
was used across 2 NVIDIA 1080 Ti graphics cards.  During the
voxelisation, random scale augmentation was applied. Applying scale
augmentation to a 3D volume on the fly, is very CPU intensive and
slows down the training procedure too much. During training,
augmentation to the input image was applied. This on-the-fly
augmentation included colour channel scaling, random translation and
random horizontal flipping.

\section{Architecture}

In the following subsections, we introduce the several architectural
options we have explored as extensions to~\cite{jackson2017vrn}. Our
first network is the same as the one used in~\cite{jackson2017vrn}, referred
to as \textit{VRN - Guided}, which establishes our baseline. This
network employs two Encoder-Decoder (``hourglass'') networks in a
stack. We follow a similar design, aside for the changes described in
this section. All of our architectures were trained with the same loss
function as in~\cite{jackson2017vrn}:

\begin{equation}
  l_{1} = \sum\limits_{w=1}^{W} \sum\limits_{h=1}^{H}\sum\limits_{d=1}^{D}[V_{whd}\log \widehat{V}_{whd}+(1-V_{whd})\log(1-\widehat{V}
_{whd})],
\end{equation}
where $\widehat{V}_{whd}$ is the corresponding sigmoid output at voxel
$\{w,h,d\}$.

\subsection{\textit{Ours - Multistack}}

\begin{figure}
  \includegraphics[width=\linewidth]{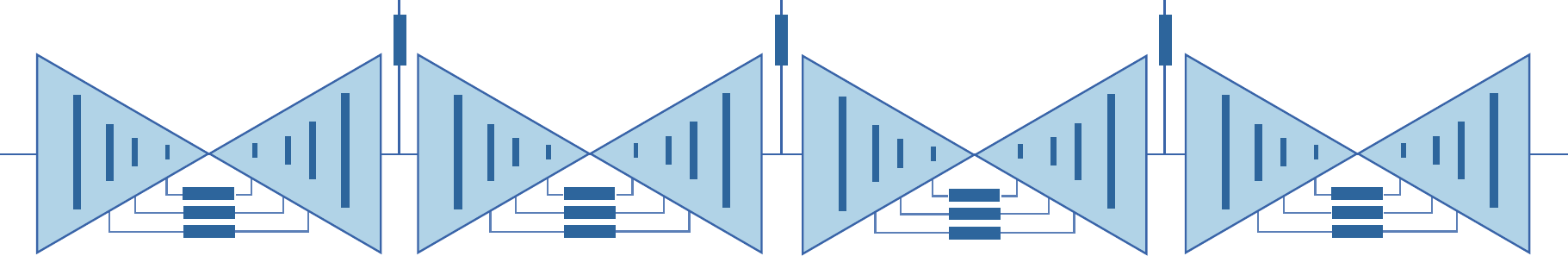}
  \caption{The \textit{Ours - Multistack} network. Dark blue boxes
    represent residual modules. Each Encoder-Decoder module has its
    own loss, while still passing features to the next module.}
  \label{fig:multistack}
\end{figure}

This network makes the following changes to the \textit{VRN - Guided}
baseline network. We half the number of residual modules from four to
two. In doing so, we also halved the memory requirements, allowing us
to increase the number of hourglass modules in the stack, from two to
four. Next, we replace the original residual module used in
\textit{VRN - Guided} with the multi-scale residual module proposed
in~\cite{bulat2017binarized}. We also show the performance improvement
from introducing just this component in the results section. Finally,
we introduce supervision after each hourglass module. We therefore
have four losses. Each hourglass module forks to provide features for
the next hourglass, and to regress the volumetric representation. The
performance after each hourglass improves. We found that there was no
benefit to adding more than four hourglass networks as the performance
just fluctuates as more are added. This network is depicted in
Figure~\ref{fig:multistack}.

\subsection{\textit{Ours - Image Only}} % multistack-nolandmarks

Our standard network (\textit{Ours - Multistack}) is trained to
receive guidance from the landmarks, while also using useful
information from the images. With this network, we try to measure the
impact of training with just images, while keeping the architecture
identical. We call this network \textit{Ours - Image Only}. We expect
that the performance of this network be significantly lower than when
guidance from the human pose is also provided. 

\subsection{\textit{Ours - Landmarks Only}} % vrn-noimg

Many methods, such
as~\cite{bogo2016smplify,ramakrishna2012reconstructing}, use only the
landmarks as input during training and inference. Hence, it is an
interesting investigation to measure the performance of our method
when only landmarks are provided, without the image. As such, we
trained \textit{Ours - Landmarks Only}. However, using only landmarks
to fit a shape model results in generic appearing fittings. Provided
high quality training data is available, our method can regress these
fine details and match the body shape when also provided with the
image.

\subsection{\textit{Ours - Mask Only}}

Our method does not rely on a segmentation mask, as is the case
in~\cite{sigal2008combined}. However, there is no reason why our
method cannot reconstruct 3D geometry from a single segmentation mask,
or silhouette. To show this, we train another network, \textit{Ours -
  Segmentation Mask} which accepts only a single channel, containing
the mask of the target object. From this, the network reconstructs the
3D geometry in the same way. Once again, this network has an identical
configuration to \textit{Ours - Multistack}, apart from the first
layer receiving a different number of inputs. We expect this network
to perform quite well since the segmentation mask we are providing to
the network is the projection of the target volume.

\subsection{\textit{Ours - 3D Convolution}}

\begin{figure}
  \centering
  \includegraphics[width=0.3\linewidth]{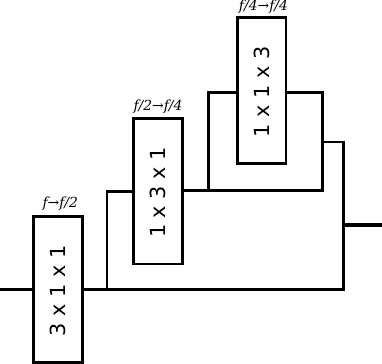}
  \caption{A ``flat'' volumetric residual block}
  \label{fig:flat_vol_residual}
\end{figure}

While volumetric CNNs can likely outdo a spatial network in terms of
performance, on this task, the memory requirements are much higher
than that of a spatial CNN. So much so, that employing volumetric CNNs
at a suitable resolution is not currently possible. However, we were
interested to test a compromise between the two and train a volumetric
CNN where the filters are flat. More concretely, where $f$ is the
number of features, our filters had sizes $f\times 3\times 1\times 1$,
$f\times 1\times 3\times 1$ or $f\times 1\times 1\times 3$. These were
combined into a flat volumetric residual module, as shown in
Figure~\ref{fig:flat_vol_residual}, heavily inspired
by~\cite{qiu2017pseudo}. This network also takes as input the image
with corresponding landmarks. To provide a fair comparison with the
other methods, we match the number of floating point operations of
this network to \textit{Ours - Multistack} by reducing the number of
parameters (which also allows the network to fit into memory).

\section{Results}

\begin{figure}
  \centering
\includegraphics[width=0.2\linewidth]{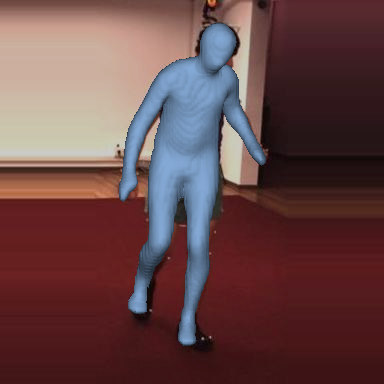}
\includegraphics[width=0.2\linewidth]{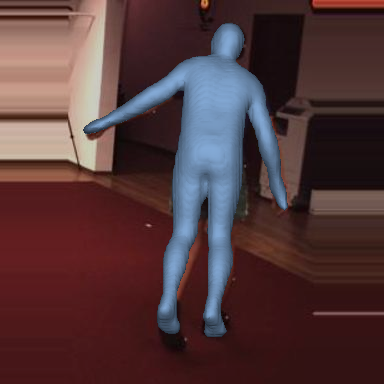}
\includegraphics[width=0.2\linewidth]{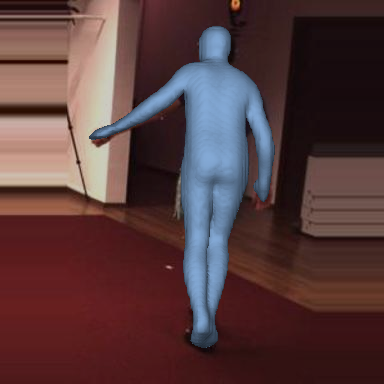}
\includegraphics[width=0.2\linewidth]{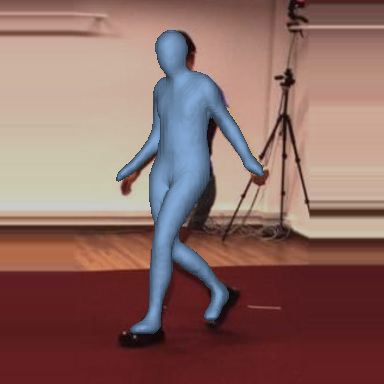} \\
\includegraphics[width=0.2\linewidth]{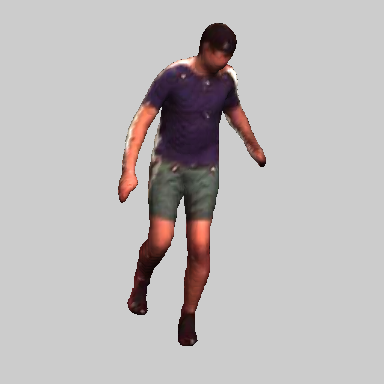}
\includegraphics[width=0.2\linewidth]{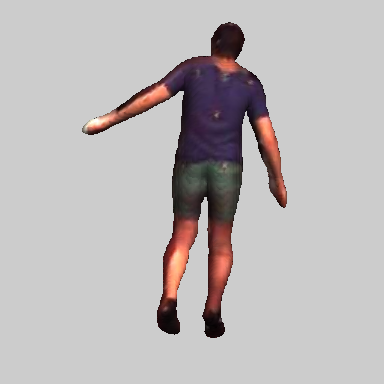}
\includegraphics[width=0.2\linewidth]{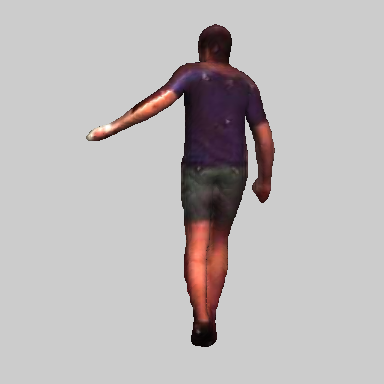}
\includegraphics[width=0.2\linewidth]{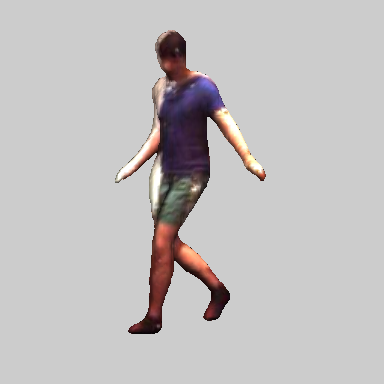} \\ [0.5em]

\includegraphics[width=0.2\linewidth]{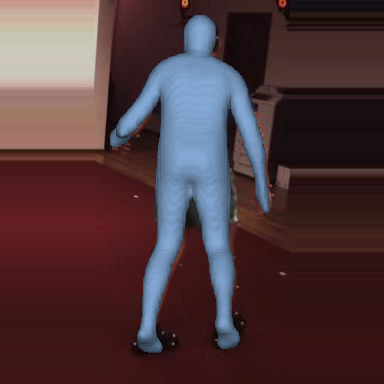}
\includegraphics[width=0.2\linewidth]{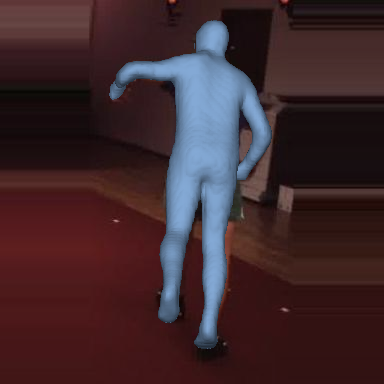}
\includegraphics[width=0.2\linewidth]{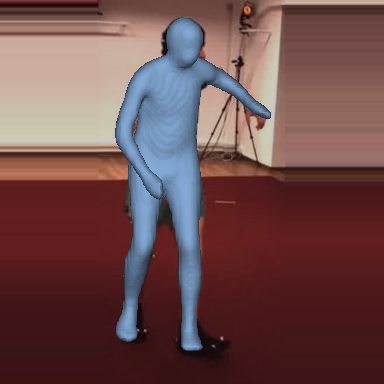}
\includegraphics[width=0.2\linewidth]{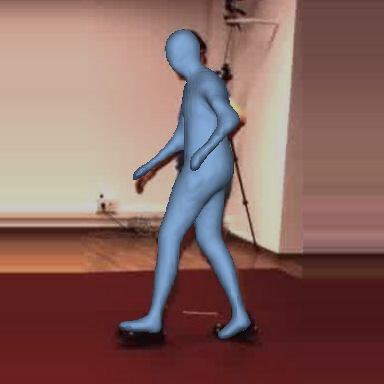} \\
\includegraphics[width=0.2\linewidth]{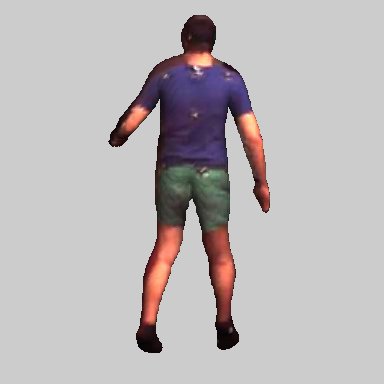}
\includegraphics[width=0.2\linewidth]{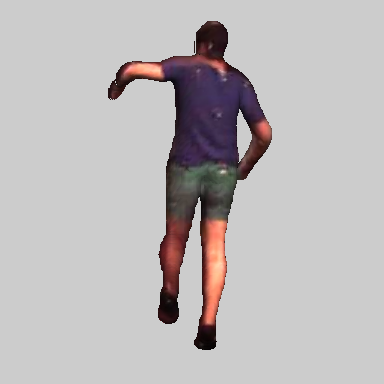}
\includegraphics[width=0.2\linewidth]{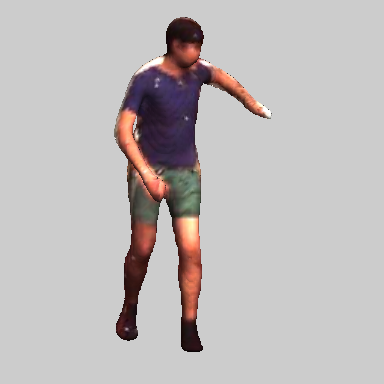}
\includegraphics[width=0.2\linewidth]{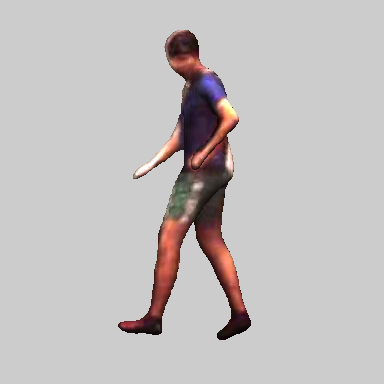}
\caption{Visual results from our main network, \textit{Ours -
    Multistack}, on a test split of Human3.6m~\cite{h36m_pami}. These
  results demonstrate VRN's ability to deal with large and complex
  pose. We also show the reconstructions with the texture projected
  onto them.}
\label{fig:visres_ours}
\end{figure}

In this section we will give an overview of the performance of the
architectures we have described above. For each network, we give our
results as an Intersection over Union (IoU) score, which is defined as
the number of intersecting set voxels over the number voxels set in
either volume. These numeric results may be found in
Table~\ref{tab:results}. We will discuss these results in more detail
in the proceeding paragraphs.

We show visual results for \textit{Ours - Multistack} in
Figure~\ref{fig:visres_ours}. The quantitative results suggest that
the changes we made to the baseline network \textit{VRN - Guided}
helped quite significantly, offering a performance increase of over
4\% in terms of IoU. From this performance improvement, more than 2\%
was due to using the residual module proposed
in~\cite{bulat2017binarized}, this can be seen from the results for
\textit{Ours - Old Residual}. As our data is generated by
SMPlify~\cite{bogo2016smplify}, we are unable to provide a
quantitative comparison with this method.

As expected, removing either the landmarks \textit{or} the image
reduces performance. The best performance is attainable by providing
the network with both the image and landmarks, as seen quantitatively
between \textit{Ours - Multistack}, \textit{Ours - Landmarks Only} and
\textit{Ours - Image Only}. Also unsurprisingly, landmarks alone
offers better performance than the image alone. This is true at least
in this case, as the groundtruth model has no detail. We also show
performance where only the segmentation mask is provided to the
network (this is not provided in the case of \textit{Ours -
  Multiststack}). These results are labelled \textit{Ours - Mask
  Only}. We expected this network to perform better than the landmarks
or image only networks, as the mask we provided was a direct 2D
projection of the target volume.

\begin{table}
  \caption{Numerical performances of our proposed method and additional
  architectural experiments, all on data generated using SMPLify.}
  \label{tab:results}
  \centering
  \begin{tabular}{|l||c|c|}
    \hline
    \textbf{Method} & \textbf{IoU @ Epoch 30}  & \textbf{IoU @ Epoch 60} \\
    \hline\hline
    \textit{VRN - Guided}  (Baseline)  & 61.6\% & 63.9\% \\
    \hline
    \textit{Ours - Multistack}         & 61.1\% & \textbf{68.3\%} \\
    \textit{Ours - Old Residual}       & 60.5\% & 66.1\% \\
    \hline
    \textit{Ours - Landmarks Only}     & 58.6\% & 61.0\%  \\
    \textit{Ours - Image Only}         & 46.8\% & 48.3\% \\
    \textit{Ours - Mask Only}          & 52.8\% & 53.0\%\\
    \hline
    \textit{Ours - 3D Convolution}     & 57.3\% & 61.6\%\\

    \hline
\end{tabular}
\end{table}

\paragraph{Notes on Performance.} A single forward pass through our
network takes approximately 200ms on an NVIDIA 1080 Ti GPU. This
produces the volumetric representation. Surface extraction introduces
200-600ms overhead depending on the implementation used. Significantly
higher performance may be achieved with smaller volumes, but this will
result in a lower level of detail. Training typically takes about two
days.

\section{High Quality Training Data}
\label{sec:detailed}

\begin{figure}
  \centering
\includegraphics[width=0.15\linewidth]{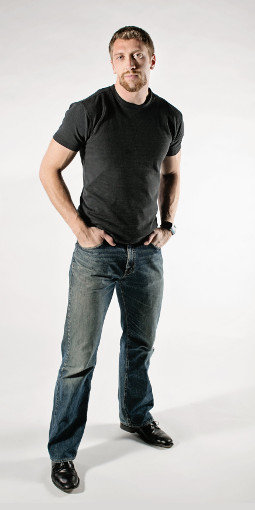}
\includegraphics[width=0.15\linewidth]{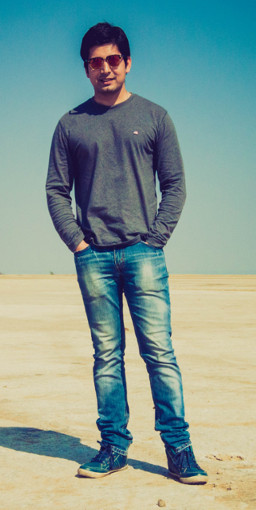}
\includegraphics[width=0.15\linewidth]{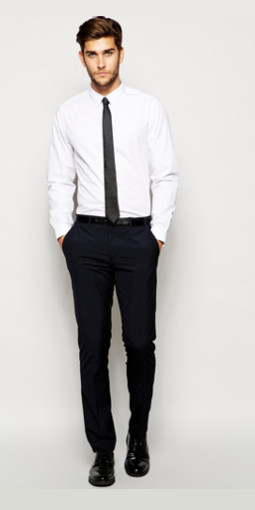}
\includegraphics[width=0.15\linewidth]{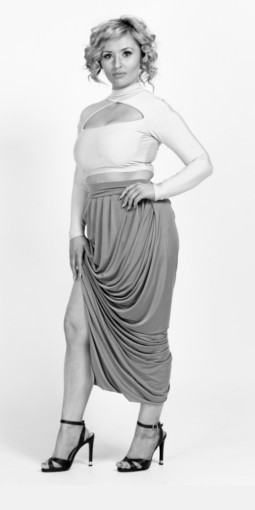}
\includegraphics[width=0.15\linewidth]{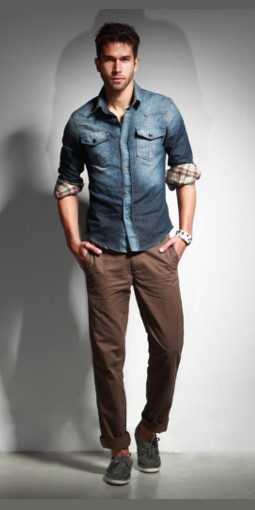}
\includegraphics[width=0.15\linewidth]{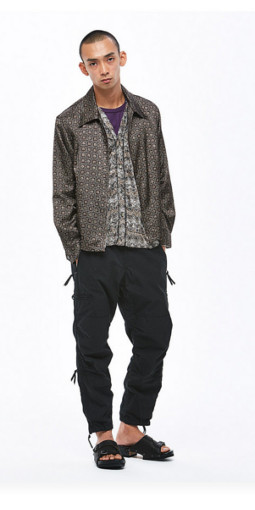}
    \\ [0.2em]
\includegraphics[width=0.15\linewidth]{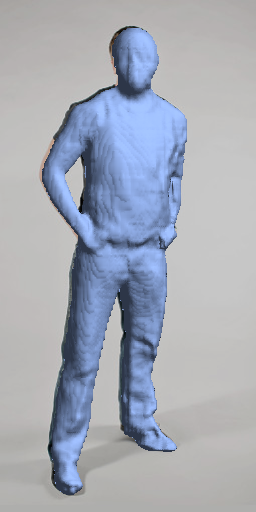}
\includegraphics[width=0.15\linewidth]{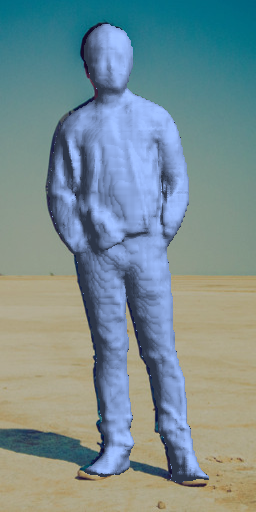}
\includegraphics[width=0.15\linewidth]{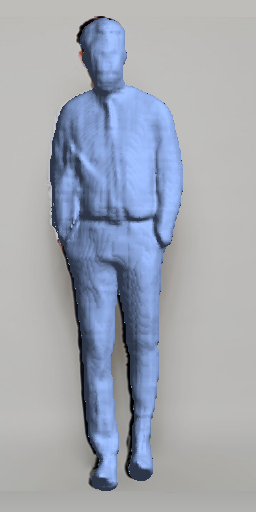}
\includegraphics[width=0.15\linewidth]{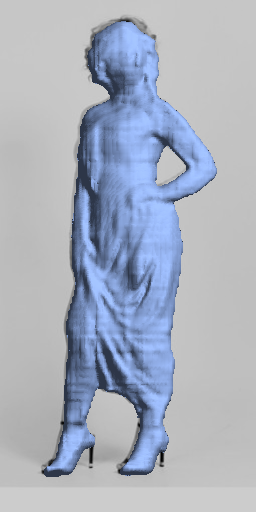}
\includegraphics[width=0.15\linewidth]{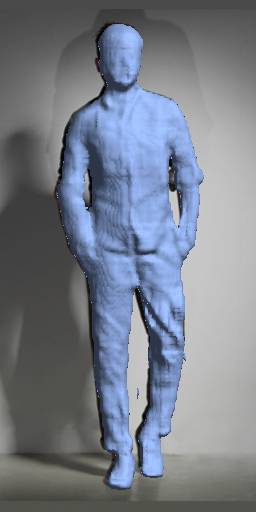}
\includegraphics[width=0.15\linewidth]{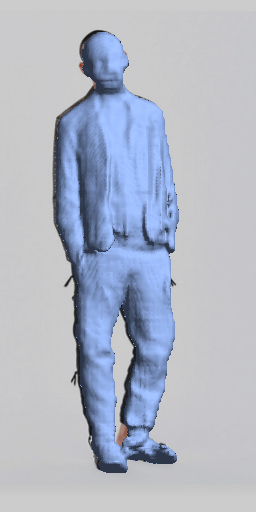}
    \\ [0.2em]
\includegraphics[width=0.15\linewidth]{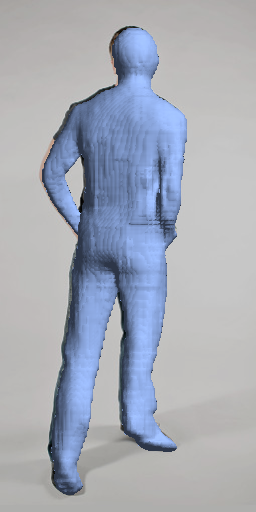}
\includegraphics[width=0.15\linewidth]{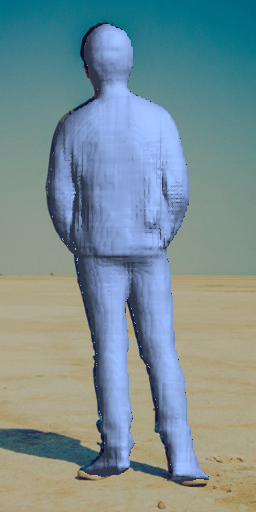}
\includegraphics[width=0.15\linewidth]{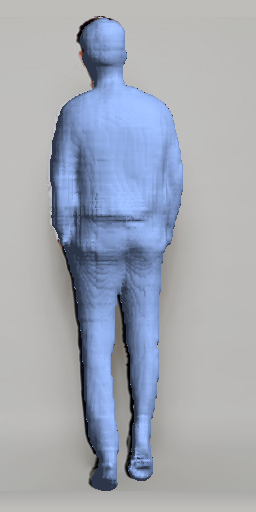}
\includegraphics[width=0.15\linewidth]{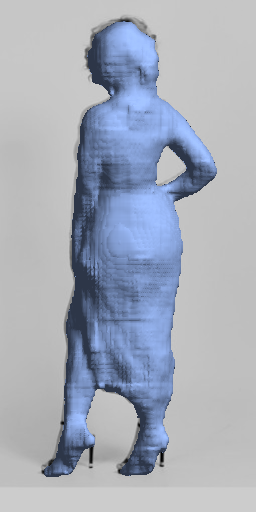}
\includegraphics[width=0.15\linewidth]{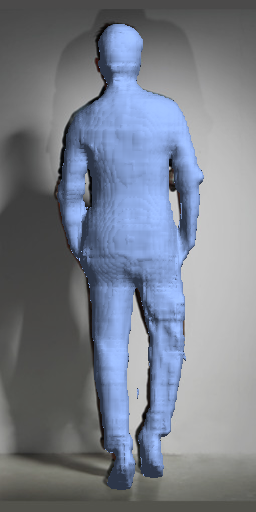}
\includegraphics[width=0.15\linewidth]{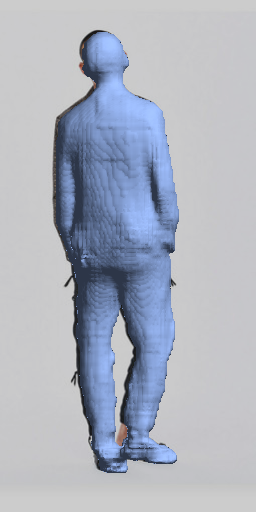}
    \\ [0.2em]
\includegraphics[width=0.15\linewidth]{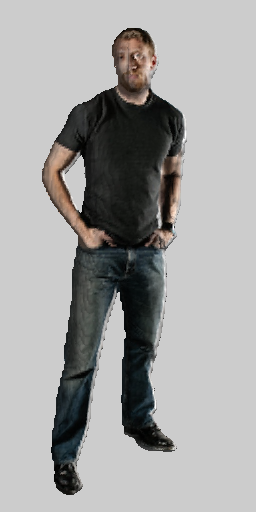}
\includegraphics[width=0.15\linewidth]{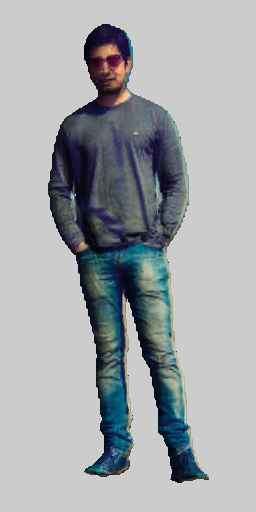}
\includegraphics[width=0.15\linewidth]{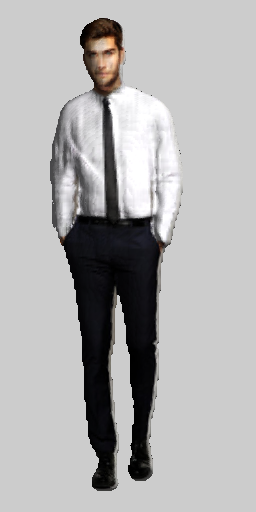}
\includegraphics[width=0.15\linewidth]{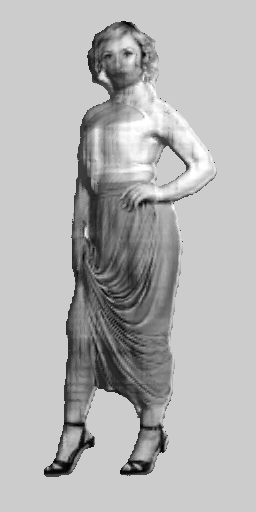}
\includegraphics[width=0.15\linewidth]{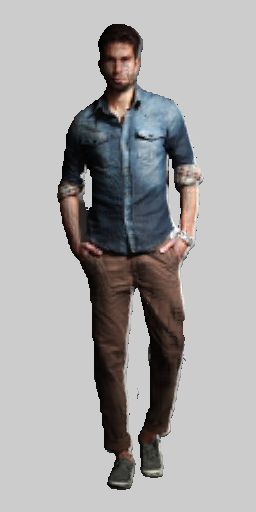}
\includegraphics[width=0.15\linewidth]{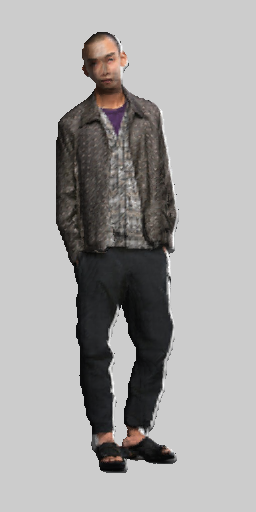}
\caption{Example 3D reconstructions from the web (Creative Commons)
  using our method trained with high quality training data. The first
  row shows the input image, the second shows the 3D reconstruction
  from the front, and the third row shows the 3D reconstruction when
  views from behind (i.e. the hallucinated side, in the case of these
  images). The final row shows the frontal reconstruction with the
  projected texture. These results show that VRN is capable of
  regressing finer details.}
  \label{fig:detailed}
\end{figure}

In the previous section, we showed that our method can reconstruct
bodies of very large pose. However, due to the dataset we trained on,
we are only able to regress the coarse geometry without any
detail. Detailed 3D reconstruction was also not demonstrated in the
case of faces in~\cite{jackson2017vrn}, which was also due to the lack
of a detailed dataset. Hence, in this section, we demonstrate that VRN
\textit{is} capable of regressing details when a high quality dataset
is provided. For this experiment, we use our best performing network
\textit{Ours - Multistack}.

Our dataset consists of highly detailed 3D scans from 40 participants,
4 of which were reserved for quantitative testing, but all of which
are quite restricted in terms of pose. Only one scan per participant
was available. These models do not have a corresponding image which is
aligned with the model. As such, we rendered and voxelised these
models under a large variety of different lighting conditions, scales
and views to create our training set consisting of approximately
20,000 samples which are spatially aligned.  The voxelisation was
performed at a resolution of $128\times 256\times 96$, which
efficiently encapsulates the poses found in the dataset. As in our
previous experiment, the Z alignment was performed by the mean Z
component. Unfortunately we are not able to publicly release this
dataset.

\subsection{Performance}

The four models which we reserved for testing were also rendered and
voxelised in the same way as above, to produce 60 testing images. Our
method reconstructs these with an IoU of 78\%. This is significantly
higher than the reconstructions in our previous experiment. This is
likely due to the better spatial alignment between the training images
and target. Additionally, we show qualitative results on real-world
images taken from the web~\footnote{These images are licensed under
  Creative Commons. Attribution, where required, will be provided on
  our website.}. These reconstructions can be found in
Figure~\ref{fig:detailed}. We show the backsides of these
reconstructions, which demonstrate the networks ability to reconstruct
the self-occluded body parts. Finer details can be seen in the
wrinkles of clothing. As our method was trained on synthetic data, we
believe there may be some performance degradation on real-world
images. Additionally, several of the poses found in the
reconstructions in Figure~\ref{fig:detailed} are not found in the 36
training samples. This suggests that VRN is somewhat tolerant to
previously unseen poses.

% https://www.flickr.com/photos/albertscherbatsky/16061100679/
% https://www.publicdomainpictures.net/en/view-image.php?image=208575&picture=businesswoman-with-a-bag
% https://www.flickr.com/photos/maylee213/7266792026/
% https://pixabay.com/en/model-studio-fashion-female-woman-1802143/
% https://pxhere.com/en/photo/691305
% https://pxhere.com/en/photo/1192855
% https://pixabay.com/en/full-body-shot-photography-people-1584734/
% https://www.flickr.com/photos/73034986@N00/32715962442

\section{Conclusion}

In this work we have shown that using Volumetric Regression Networks,
as described in~\cite{jackson2017vrn}, for the task of 3D
reconstruction, is not restricted to the simpler task of face
reconstruction. Nor is it a limiting factor in terms of detail,
despite the small size of the volumes we are working with. We have
proposed several improvements to the original VRN which improve the
performance quite substantially. Finally, we have shown, by using two
different datasets, that VRN can regress both unusual poses (in
networks trained on Human3.6m), and high levels of detail (in the case
of our private but detailed dataset). We believe that given a large
enough dataset containing many pose variations, and high levels of
detail, the network will be capable of large pose 3D human
reconstruction, while also capturing fine details, from a single
image.

\section{Acknowledgements}

Aaron Jackson is funded by a PhD scholarship from the University of
Nottingham. Thank you to Chris Manafas and his team at 2B3D for
providing data for the experiments. We are grateful for access to the
University of Nottingham High Performance Computing Facility, which
was used for data voxelisation.

% \clearpage

\bibliographystyle{splncs}
\bibliography{egbib}
\end{document}